\documentclass[10pt,logo,copyright]{nvidiatechreport}
\usepackage[authoryear,round]{natbib}

\usepackage[utf8]{inputenc} % allow utf-8 input
\usepackage[T1]{fontenc}    % use 8-bit T1 fonts

\usepackage{parskip}        % no paragraph indents
\usepackage{url}            % simple URL typesetting
\usepackage{booktabs}       % professional-quality tables
\usepackage{amsfonts}       % blackboard math symbols
\usepackage{nicefrac}       % compact symbols for 1/2, etc.
\usepackage{microtype}      % microtypography
\usepackage{xcolor}         % colors
\usepackage[dvipsnames]{xcolor} % more color names
\usepackage{graphicx}
\usepackage{animate}        % for 360 video in teaser
\usepackage{subcaption}
\usepackage{tabularx}
\usepackage{makecell}
\usepackage{adjustbox}
\usepackage{setspace}
\newcolumntype{M}[1]{>{\centering\arraybackslash}m{#1}}
\usepackage{float}
\usepackage{tikz}
\usetikzlibrary{positioning,shapes,arrows}
\usepackage{amsmath,amsfonts,bm, bbm,leftindex}
\usepackage{multirow}
\usepackage{comment}
\usepackage{gensymb}
\usepackage{lipsum}
\usetikzlibrary{arrows.meta, positioning, fit}
\usepackage[para]{threeparttable}
\usepackage{tikz}
\usetikzlibrary{tikzmark}

\usepackage[most]{tcolorbox}
\usepackage{fancyvrb}
\usepackage{fvextra}
\usepackage{dashrule}

\usepackage{algorithm}
\usepackage{algpseudocode}

\usepackage{multirow}
\usepackage{rotating}
\usepackage{dsfont}
\usepackage{pifont}
\usepackage{wrapfig}

\usepackage[capitalize]{cleveref}
\crefname{section}{Sec.}{Secs.}
\Crefname{section}{Section}{Sections}
\Crefname{table}{Table}{Tables}
\crefname{table}{Tab.}{Tabs.}
\crefname{equation}{Eq.}{Eqs.}
\crefname{algorithm}{Alg.}{Algs.}
\Crefname{algorithm}{Algorithm}{Algorithms}

\newcommand{\NA}{\textcolor{gray!50}{\ding{55}}} % Not Applicable
\definecolor{Red7}{RGB}{240, 62, 62}
\definecolor{Pink7}{RGB}{214, 51, 108}
\definecolor{Grape7}{RGB}{174, 62, 201}
\definecolor{Violet7}{RGB}{112, 72, 232}
\definecolor{Indigo7}{RGB}{66, 99, 235}
\definecolor{Blue7}{RGB}{28, 126, 214}
\definecolor{Cyan7}{RGB}{16, 152, 173}
\definecolor{Teal7}{RGB}{12, 166, 120}
\definecolor{Green7}{RGB}{55, 178, 77}
\definecolor{Lime7}{RGB}{116, 184, 22}
\definecolor{Yellow7}{RGB}{245, 159, 0}
\definecolor{Orange7}{RGB}{247, 103, 7}

\definecolor{HistoryMarker}{RGB}{213, 234, 253}
\definecolor{HistoryCompressor}{RGB}{240, 218, 249}

\definecolor{maskgreen}{rgb}{0.639,0.984,0.639}
\definecolor{maskpink}{rgb}{1,0.803,0.973}
\definecolor{maskorange}{rgb}{0.988,0.827,0.635}
\definecolor{maskblue}{rgb}{0.624,0.804,1}
\definecolor{nvgreen}{rgb}{0.462,0.725,0}
\definecolor{brightlavender}{rgb}{0.75, 0.58, 0.89}

\newcommand{\model}{AUSM}

\renewcommand{\copyrightext}{%
    \footerfont 
    $^*$ Equal contribution. $^\dagger$ Work partially done during internship at NVIDIA. $^\ddagger$ Corresponding authors. \\
    \textcopyright\, \the\year{} NVIDIA. All rights reserved.
}

\title{\centering Autoregressive Universal Video Segmentation Model}

\author{ \centering
    {Miran Heo}$^{*,\dagger,1,2}$\quad {Sukjun Hwang}$^{*,3}$\quad {Min-Hung Chen}$^1$ \quad {Yu-Chiang Frank Wang}$^{1,4}$

    \vspace{-3mm}
    {Albert Gu}$^{3}$\quad {Seon Joo Kim}$^{\ddagger, 2}$\quad {Ryo Hachiuma}$^{\ddagger,1}$ \\
    \vspace{1mm}
    \normalsize \textsuperscript{1} NVIDIA \quad
    \normalsize \textsuperscript{2} Yonsei University \quad
    \normalsize \textsuperscript{3} Carnegie Mellon University \quad
    \normalsize \textsuperscript{4} National Taiwan University
}

\begin{document}

\maketitle

\begin{abstract}
\label{sec:abstract}
Recent video foundation models such as SAM2 excel at prompted video segmentation by treating masks as a general-purpose primitive.
However, many real-world settings require unprompted segmentation that aims to detect and track all objects in a video without external cues, leaving today's landscape fragmented across task-specific models and pipelines.
We recast streaming video segmentation as sequential mask prediction, analogous to language modeling, and introduce the Autoregressive Universal Segmentation Model (AUSM), a single architecture that unifies both prompted and unprompted video segmentation.
Built on recent state-space models, AUSM maintains a fixed-size spatial state and scales to video streams of arbitrary length.
Furthermore, all components of AUSM are designed for parallel training across frames, yielding substantial speedups over iterative training.
On standard benchmarks (DAVIS17, YouTube-VOS 2018 \& 2019, MOSE, YouTube-VIS 2019 \& 2021, and OVIS) AUSM outperforms prior universal streaming video segmentation methods and achieves up to 2.5$\times$ faster training on 16-frame sequences.
\end{abstract}
\abscontent
\section{Introduction}
\label{sec:intro}

Language and video are both sequential modalities that arrive as streams, yet their modeling trajectories have diverged. In language, decoder-only large language models (LLMs) show that a single scalable architecture trained on massive corpora can subsume diverse tasks. Video perception would benefit from the same unification: annotations are fragmented across tasks while video data is expensive to curate, so a unified model could amortize supervision and deployment. We focus on streaming video segmentation, and an ideal universal model should: (i) accommodate a broad set of tasks, (ii) preserve fine-grained spatio-temporal details from past inputs, (iii) support inference over long videos, and (iv) enable training that scales efficiently with sequence length. Despite the structural parallels to language, current practice in video remains partitioned into task-specific architectures and training protocols, and no existing approach simultaneously meets all four criteria.

We categorize streaming video segmentation into two regimes: prompted and unprompted. Prompted video segmentation, exemplified by video object segmentation (VOS), takes an initial human cue (\emph{e.g.,} mask, box, point, or text) and propagates specified targets over time; it is effective for user-interactive editing, but does not naturally handle the emergence of novel instances without re-prompting (\emph{e.g.,} in autonomous driving). Unprompted video segmentation -- comprising video instance and panoptic segmentation (VIS/VPS) -- aims to detect and track all instances of predefined categories throughout a video. Many detect-then-track pipelines \citep{VIS, MinVIS, IDOL} process frames independently and associate \emph{post hoc}; even approaches that leverage past information \citep{IDOL, CTVIS, VISAGE} compress each instance into a few vectors, preserving history mainly for identity association and discarding fine-grained spatio-temporal details needed for detection. Recent attempts at universal models~\citep{UNINEXT, UniVS, TarViS} retrofit VOS into VIS-style architectures by encoding an instance as a heavily compressed token, which yields noticeable performance drops on VOS. Furthermore, to our knowledge, existing training frameworks for video segmentation lack LLM-style parallelized training, limiting scalability with respect to sequence length.

In this paper, we present an autoregressive universal video segmentation model, \textbf{AUSM}, that unifies both prompted and unprompted video segmentation tasks while scaling to long video settings.

\textbf{Unified Structure.} First, we identify a key connection between the autoregressive pipeline of decoder-only LLMs and streaming video perception: \textbf{next-word prediction as next-frame mask prediction}. Based on this perspective, we present a unification of prompted and unprompted video segmentation and design a universal model that achieves state-of-the-art performance among online universal methods on seven benchmarks comprising both VOS and VIS using shared weights.

\textbf{Architecture.} AUSM is designed with components specialized for streaming videos, \textbf{History Marker} and \textbf{History Compressor}. History Marker removes abstraction of instances by leveraging Token Mark~\citep{Omni-RGPT} and dissolves segmentation masks into frame features for retrieval of instance-wise information. This effectively preserves fine-grained information, demonstrating a nearly 10\% improvement in VOS performance compared to previous unified online architectures. History Compressor then takes the output and compresses spatio-temporal information of all past frames into a single spatial state. While video segmentation models typically store fewer than ten frame features~\citep{STM, SAM2}, our design makes processing arbitrarily long streams feasible.

\textbf{Training Acceleration.} AUSM supports \textbf{parallel training}, a critical property of the building blocks~\citep{Transformer, Mamba} used in decoder-only LLMs for extending to long sequences. Compared to existing frameworks that recurrently process frames during training~\citep{RoCoVIS, AOT, SAM2}, our training pipeline shows significant speedups.

Finally, we evaluate AUSM across a diverse set of benchmarks spanning both prompted and unprompted video segmentation: DAVIS 2017~\citep{DAVIS}, YouTube-VOS 18\&19~\citep{YTVOS}, MOSE~\citep{MOSE}, YouTube-VIS 19\&21~\citep{VIS}, and OVIS~\citep{OVIS}.
AUSM delivers strong performance across all benchmarks, outperforming previous online universal video segmentation models~\citep{UniVS, UNINEXT}.
Importantly, these results are achieved \emph{without relying on FIFO memory buffers}~\citep{STM, SAM}, highlighting the efficiency and scalability of our autoregressive design.
Furthermore, parallel training becomes faster compared to recurrent training frameworks as the sequence length increases, achieving up to $2.5\times$ faster training than iterative baselines with 16-frame sequences.

\section{Autoregressive Universal Video Segmentation Model (AUSM)}

\begin{figure}[t]
\begin{center}
\includegraphics[width=\linewidth]{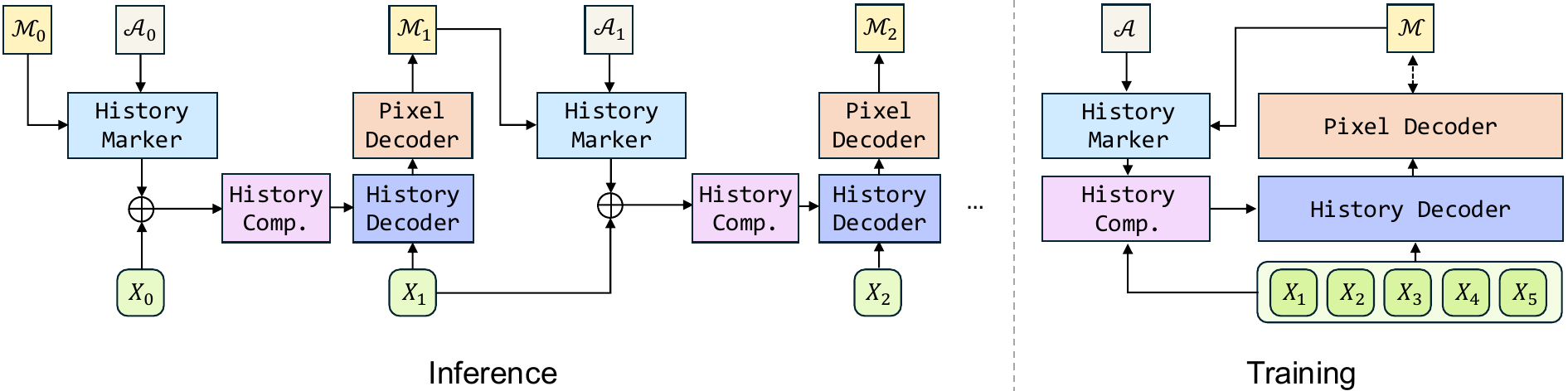}
\end{center}
% \vspace{-2mm}
\caption{
High-level overview of \model{} during training and inference. 
\textbf{Left:} At inference time, \model{} processes frames in a recurrent manner with constant decoding time per frame. Rather than maintaining an explicit memory buffer, temporal history is compressed via the Mamba layer within the History Compressor module and passed through time. 
\textbf{Right:} During training, a parallel formulation is used to jointly optimize over multiple frames, enabling efficient and scalable learning.
}

\label{fig:training_inference}
\end{figure}

In \cref{sec: method_autoregressive}, we first formalize video segmentation within an autoregressive framework inspired by LLMs, which provides a seamless unification of prompted and unprompted video segmentation tasks.
In \cref{sec:architecture}, we then introduce the architectural components and overall algorithm of AUSM using the recurrent form (\cref{fig:training_inference}-Left).
Finally, in \cref{sec: parallel}, we demonstrate how each component of AUSM is designed for parallel training (\cref{fig:training_inference}-Right).

\subsection{Video Segmentation as Language Modeling}
\label{sec: method_autoregressive}

Video segmentation encompasses a family of tasks~\citep{VIS, DAVIS, VPS, MOTS, Cityscapes_VSS} with varying objectives but a shared core definition. Given a video consisting of $T$ frames, $\left(\mathcal{I}_1, \ldots, \mathcal{I}_T\right)$, where each frame $\mathcal{I}_t \in \mathbb{R}^{H \times W \times 3}$ is an RGB image with spatial resolution $H \times W$, the goal is to produce $\left(\hat{y}_1, \ldots, \hat{y}_T \right)$ that predicts a sequence of segmentation ground truth $Y = \left(y_1, \ldots, y_T\right)$.
Each $y_t$ represents ground truth at time $t$ and is defined as a set $\{(c^i_t, m^i_t)\}_{i=1}^{N^\text{gt}}$, where $c^i_t \in \{1, \ldots, K\}$ denotes the class label of the $i$-th object, $m^i_t \in \{0,1\}^{H \times W}$ is its segmentation mask, and $N^\text{gt}$ is the number of foreground objects.
Similarly, each $\hat{y}_t=\{(\hat{c}_t^i, \hat{m}_t^i)\}_{i=1}^{N^\text{det}}$ represents predictions at time $t$, where $\hat{c} \in\{1, \ldots, K, \texttt{bg}\}$ denotes the class label (\texttt{bg} indicates background), $\hat{m} \in \{0,1\}^{H \times W}$ is the predicted mask, and $N^\text{det}$ is the number of object queries.

We observe that video segmentation can be reformulated within an autoregressive framework used in modern LLMs~\citep{GPT3, Llama3}.
Language models generate sequences by conditioning each token on all previous tokens:
\begin{equation}
\label{eq: autoregressive}
P(y_{1:T}) = \prod_{t=1}^{T} P(y_t \mid y_{<t}) .
\end{equation}
The sequential nature of language naturally aligns with this autoregressive formulation, which provides an elegant unification of different tasks in language into a single universal architecture~\citep{GPT2}.
Similarly, the segmentation of a streaming video can be expressed as
\begin{equation}
    P(y_{1:T} \mid \mathcal{I}_{1:T}) = \prod_{t=1}^{T} P(y_t \mid y_0, y_{<t}, \mathcal{I}_{\leq t}) ,
\end{equation}
which explicitly models the dependence of each frame's segmentation $y_t$ on the current frame $\mathcal{I}_t$, all previous frames $\mathcal{I}_{<t}$, previous segmentations $y_{<t}$, and potentially an initial prompt $y_0$~\citep{DAVIS}.
This formulation covers all video segmentation tasks: prompted video segmentation begins with an initial human prompt $y_0=m_0$, while no initial prompt is given for unprompted video segmentation ($y_0 = \varnothing$)~\citep{VIS, VPS}.

\subsection{\model{} Architecture}
\label{sec:architecture}

The design of \model{} allows sophisticated conditioning on past history while maintaining constant memory to process arbitrarily long video sequences.
As shown in \cref{alg: recurrent}, AUSM uses a pool of object queries $\mathcal{V} \in \mathbb{R}^{N^\text{det} \times D}$ and keeps a set of buffer ID vectors $\mathcal{B} \in \mathbb{R}^{N^{\text{id}} \times D}$ that are essential for tracking identified instances, where $N^{\text{id}}$ is the total number of available ID vectors.
Additionally, AUSM maintains two sets during inference: $\mathcal{A}$ for storing allocated ID vectors and $\mathcal{M}$ for storing previous mask predictions.
These sets maintain a one-to-one mapping, ensuring that $\mathcal{A}_t^i$ ($i$-th element of $\mathcal{A}_t$) corresponds to $\mathcal{M}_t^i$ ($i$-th element of $\mathcal{M}_t$), thus $|\mathcal{A}_t|=|\mathcal{M}_t|$ at all times.
We assume each frame is encoded by a frame-independent backbone, yielding features $\{X_1,\ldots,X_T\}$ where $X_t = \texttt{backbone}(\mathcal{I}_t) \in \mathbb{R}^{H \times W \times D}$ for $t \in \{1,\ldots,T\}$, and $D$ is the channel dimension.
Additionally, we define $X_0 \in \mathbb{R}^{H \times W \times D}$ by spatially repeating a $D$-dimensional trainable vector.

\begin{algorithm}
\small
\caption{AUSM in Recurrent Form for Inference.}
\label{alg: recurrent}
\begin{algorithmic}[1]
\State Initialize $\mathcal{B} \in \mathbb{R}^{N^{\text{id}} \times D}$, $\mathcal{V} \in \mathbb{R}^{N^{\text{det}} \times D}$
\If{$y_0 = \varnothing$}
    \State $\mathcal{A}_0, \mathcal{M}_0 \gets \varnothing, \varnothing$
    \State $\mathcal{B}_0 \gets \mathcal{B}$
\Else
    \State $\mathcal{A}_0 \gets \texttt{Sampler}(\mathcal{B}, |y_0|)$
    \State $\mathcal{M}_0 \gets m_0$
    \State $\mathcal{B}_0 \gets \mathcal{B} \setminus \mathcal{A}_0$
\EndIf
\For{$t = 1$ to $T$}
    \State $E_t \gets X_{t-1} +\colorbox{HistoryMarker}{\texttt{HistoryMarker}}(\mathcal{A}_{t-1}, \mathcal{M}_{t-1})$
    \State $F_t \gets \colorbox{HistoryCompressor}{\texttt{HistoryCompressor}}(E_t)$
    \State $G_t \gets \colorbox{Indigo7!20}{\texttt{HistoryDecoder}}(Q=X_t, KV=F_t)$
    \State $\hat{y}_t^{\text{trk}}, \hat{y}_t^{\text{det}} \gets \colorbox{Orange7!10}{\texttt{PixelDecoder}}(Q=\texttt{concat}(\mathcal{A}_{t-1}, \mathcal{V}), KV = G_t)$
    \State $\mathcal{D} \gets \texttt{filter\_fg}(\hat{y}_t^\text{det})$
    \State $\mathcal{A}' \gets \texttt{Sampler}(\mathcal{B}_{t-1}, |\mathcal{D}|)$
    \State $\mathcal{A}_t \gets \texttt{concat}(\mathcal{A}_{t-1}, \mathcal{A}')$
    \State $\mathcal{B}_t \gets \mathcal{B}_{t-1} \setminus \mathcal{A}'$
    \State $\mathcal{M}_t \gets \texttt{concat}(\hat{m}^{\text{trk}}_t, \mathcal{D})$
\EndFor
\end{algorithmic}
\end{algorithm}

\textbf{Unification of Tasks.}
By altering the initialization of $\mathcal{A}$ and $\mathcal{M}$, AUSM handles both prompted and unprompted video segmentation.
For unprompted video segmentation, both $\mathcal{A}_0$ and $\mathcal{M}_0$ are initialized as empty sets (Line 3 in \cref{alg: recurrent}).
The initialization for prompted video segmentation involves $\texttt{Sampler}(\mathcal{B}, n)$, which uniformly samples $n$ vectors from $\mathcal{B}$ without replacement and returns them as a matrix whose rows are the sampled vectors.
Formally,
\begin{gather*}
    \texttt{Sampler}(\mathcal{B}, n) = [b_1, \ldots, b_n]^\top \in \mathbb{R}^{n \times D},\\
    \text{where } (b_1, \ldots, b_n) \sim \text{Uniform}\{(b'_1, \ldots, b'_n) \in \mathcal{B}^n : b'_i \neq b'_j, \forall i \neq j\}.
\end{gather*}
In the prompted setting, $\mathcal{A}_0 = \texttt{Sampler}(\mathcal{B}, |y_0|)$ and $\mathcal{M}_0 = m_0$, while $\mathcal{B}_0 = \mathcal{B} \setminus \mathcal{A}_0$ (Lines 6–7 in \cref{alg: recurrent}).

\textbf{\colorbox{HistoryMarker}{History Marker.}}
The vectorization of objects widely used in VIS~\citep{VITA, UniVS, GenVIS} loses spatial detail due to excessive compression of object-wise information.
In contrast, History Marker preserves fine details similarly to memory-based methods in VOS~\citep{STM}.
Specifically, it leverages Token Mark~\citep{Omni-RGPT} to dissolve instance masks into a spatial feature map, minimizing information loss~\citep{AOT, RoCoVIS}.
At time $t$, given allocated vectors $\mathcal{A}_{t-1}$ and segmentation masks $\mathcal{M}_{t-1}$, this module operates as:
\begin{equation*}
    \texttt{HistoryMarker}(\mathcal{A}_{t-1}, \mathcal{M}_{t-1}) = S_t \in \mathbb{R}^{H \times W \times D}, \quad
    S_{t}[h,w,:] = \frac{\sum_{i=1}^{|\mathcal{A}_{t-1}|} \mathcal{M}_{t-1}^{i}[h,w] \cdot \mathcal{A}_{t-1}^i}{\epsilon + \sum_{i=1}^{|\mathcal{A}_{t-1}|} \mathcal{M}_{t-1}^{i}[h,w]},
\end{equation*}
where $\epsilon$ is a small number to prevent division by zero.
The output $S_t$ is added to $X_{t-1}$ to form $E_t$, which is then passed to the History Compressor.

\begin{wrapfigure}{r}{0.35\textwidth}
  \centering
  % \vspace{-4mm}
  \includegraphics[width=0.35\textwidth]{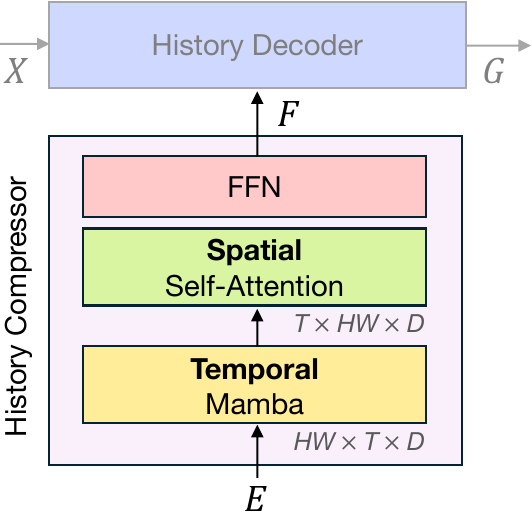}
  % \vspace{-5mm}
  \caption{\texttt{HistoryCompressor} module. Mamba encodes temporal dependencies, while self-attention captures spatial structure, enabling recurrent compression of $E$ with constant memory. 
  Specifically, the temporal mamba layer operates pixel-wise, mixing information of each pixel throughout the time dimension $T$.
  The spatial self-attention layer, by contrast, is frame-independent that fuses information spanning over $HW$ pixels.}
  \label{fig:compressor}
  \vspace{-10mm}
\end{wrapfigure}

\textbf{\colorbox{HistoryCompressor}{History Compressor.}}
This module encodes visual features from previous frames and instance-specific masks into a single spatial state, enabling inference over arbitrarily long videos with constant memory.
As shown in~\cref{fig:compressor}, each layer in the History Compressor consists of three components: Mamba~\citep{Mamba}, Self-Attention~\citep{Transformer}, and a feed-forward network.
By decomposing video features into spatial and temporal dimensions, these components operate on different axes: Mamba processes the temporal dimension while self-attention handles the spatial dimension.
We choose Mamba for the temporal dimension for two reasons: (1) videos are inherently sequential in time, aligning with SSM architectures; and (2) modeling videos is memory-intensive on GPUs because each frame contributes many tokens.
The recurrent design of Mamba enables a single spatial state that is updated each frame, eliminating the need to store spatio-temporal features from all previous frames.

\textbf{\colorbox{Indigo7!20}{History Decoder.}}
The History Decoder is a stack of Transformer decoder layers that outputs $G_t$ by taking the current-frame features $X_t$ as queries and the compressed state $F_t$ as keys and values.
This spatial feature $G_t$ has two important properties: (1) it incorporates current-frame information through the image encoder, and (2) it retains fine-grained information about objects from previous frames through the compressed state.

\textbf{\colorbox{Orange7!10}{Pixel Decoder} and Update Process.}
The Pixel Decoder follows \cite{Mask2Former}, comprising Transformer decoder layers with masked attention, and takes object queries as in Hungarian matching–based detection methods~\citep{DETR, MaskFormer, IFC}.
Specifically, the final predictions for frame $t$ are obtained using both previously allocated ID vectors $\mathcal{A}_{t-1}$ and object queries $\mathcal{V}$ as inputs, with $G_t$ providing keys and values.
Each ID vector in $\mathcal{A}_{t-1}$ tracks its corresponding object, and $\mathcal{V}$ detects instances not yet assigned to any vector in $\mathcal{A}_{t-1}$.
Therefore, this process yields two types of predictions: tracking predictions $\hat{y}_t^\text{trk}$ from $\mathcal{A}_{t-1}$ and detection predictions $\hat{y}_t^\text{det}$ from $\mathcal{V}$.

After prediction, the sets $\mathcal{A}$, $\mathcal{B}$, and $\mathcal{M}$ are updated for the next frame.
We define the following operations used in \cref{alg: recurrent}:
\begin{itemize}
    \item $\texttt{filter\_fg}(\hat{y}_t^\text{det})$: Filters predictions from $\hat{y}_t^\text{det}$ to retain only foreground objects, returning a set $\mathcal{D}$ of newly detected objects.
    \item $\texttt{concat}(A, B)$: Concatenates two sets of vectors $A$ and $B$.
\end{itemize}
The update process proceeds as follows: (1) foreground detections $\mathcal{D}$ are filtered from $\hat{y}_t^\text{det}$;
(2) $|\mathcal{D}|$ new vectors $\mathcal{A}'$ are sampled from the remaining buffer $\mathcal{B}_{t-1}$; (3) $\mathcal{A}_t$ is formed by concatenating $\mathcal{A}_{t-1}$ and $\mathcal{A}'$, while $\mathcal{B}_t$ is obtained by removing $\mathcal{A}'$ from $\mathcal{B}_{t-1}$; (4) $\mathcal{M}_{t}$ is updated by concatenating mask predictions from $\hat{y}_t^\text{trk}$ and from the newly detected objects $\mathcal{D}$.

\subsection{Parallel Training as Language Models}
\label{sec: parallel}

\begin{algorithm}
\small
\caption{AUSM in Parallel Form for Training. $A_{B:B+C}$ denotes $(A_B,\ldots,A_{B+C})$ for brevity.}
\label{alg: parallel}
\begin{algorithmic}[1]
\State Initialize $\mathcal{B} \in \mathbb{R}^{N^{\text{id}} \times D}$, $\mathcal{V} \in \mathbb{R}^{N^{\text{det}} \times D}$
\State $\mathcal{A} \gets \texttt{Sampler}(\mathcal{B}, N^\text{gt})$
\State $(y_{1:T}^\text{trk}, y_{1:T}^\text{det}), \mathcal{A}_{0:T-1}, \mathcal{M}_{0:T-1} \gets \texttt{Preprocess}(y_{1:T}, \mathcal{A})$
% \State $\mathcal{H}_{0:T-1} \gets (y_0, y_1^\text{trk} \ldots, y_{T-1}^\text{trk})$
\State $E_{1:T} \gets \left(X_{t-1} + \colorbox{HistoryMarker}{\texttt{HistoryMarker}}(\mathcal{A}_{t-1}, \mathcal{M}_{t-1})\right)_{1:T}$
\State $F_{1:T} \gets \colorbox{HistoryCompressor}{\texttt{HistoryCompressor}}(E_{1:T})$
\State $G_{1:T} \gets \colorbox{Indigo7!20}{\texttt{HistoryDecoder}}(Q=X_t, KV=F_t)_{1:T}$
\State $\left(\hat{y}_{t}^\text{trk}, \hat{y}_{t}^\text{det}\right)_{t=1}^T \gets \colorbox{Orange7!10}{\texttt{PixelDecoder}}(Q=\texttt{Concat}(\mathcal{A}_{t-1}, \mathcal{V}), KV=G_t)_{1:T}$
\State $\mathcal{L}_\text{total} = \sum_{t=1}^T \left[L^\text{trk}(y_t^\text{trk}, \hat{y}_t^\text{trk}) + L^\text{det}(y_t^\text{det}, \hat{y}_t^\text{det})\right]$
\end{algorithmic}
\end{algorithm}

Modern decoder-only language models~\cite{GPT3, Llama3} significantly benefit from parallel training by adopting building blocks (\emph{e.g.,} Transformers and SSMs) that support the teacher-forcing technique~\cite{Seq2Seq}.
However, this parallel training cannot be readily applied to existing video segmentation methods that rely on frame-by-frame propagation of outputs, such as those employing query propagation~\citep{TrackFormer, GenVIS, RoCoVIS}.
Therefore, existing frameworks train video segmentation models by recurrently processing frames, leading to severely inefficient training.

In contrast, as shown in \cref{alg: parallel}, all modules in AUSM are compatible with teacher forcing.
Therefore, AUSM supports parallel training across the temporal dimension, yielding substantial improvements in training efficiency.
The parallel training begins by sampling $N^\text{gt}$ vectors from the buffer $\mathcal{B}$ to form the set $\mathcal{A}$, with a one-to-one mapping to ground-truth instances.
A critical component is the \texttt{Preprocess} function, which prepares $y^\text{trk}_{1:T}$, $y^\text{det}_{1:T}$, $\mathcal{M}_{0:T-1}$, and $\mathcal{A}_{0:T-1}$.

\begin{wrapfigure}{r}{0.55\textwidth}
  \centering
  \vspace{-7.5mm}
  \includegraphics[width=0.55\textwidth]{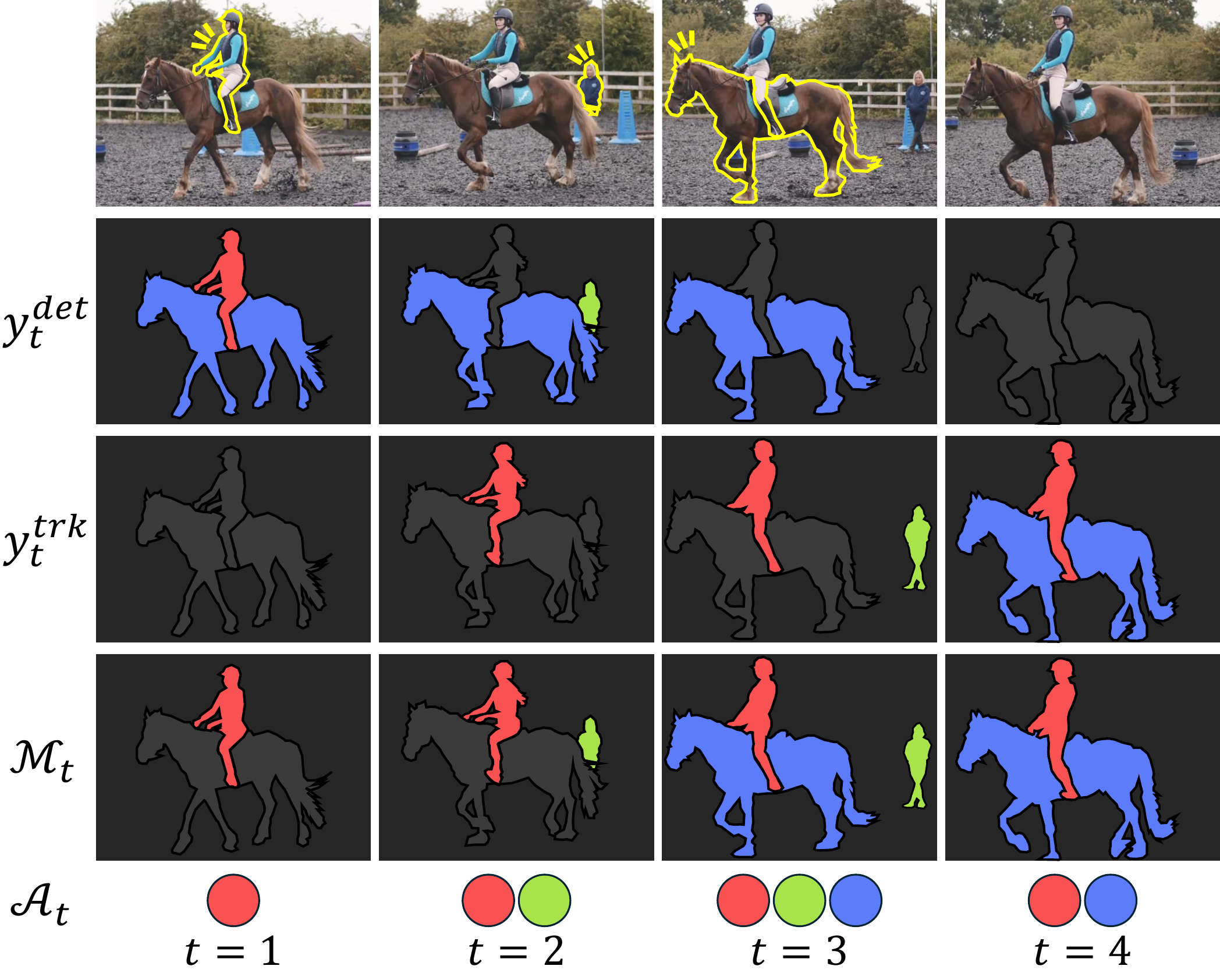}
  \vspace{-6mm}
  \caption{Schematic of \texttt{Preprocess}. The video contains three ground-truth instances: \texttt{person1} (the person riding the horse), \texttt{person2} (the person standing), and \texttt{horse}. Each instance is matched with a vector from $\mathcal{A}$, represented by colored circles: red for \texttt{person1}, green for \texttt{person2}, and blue for \texttt{horse}. The highlighted contours indicate the randomly sampled timesteps ($t_\text{sample}^i$) for each instance.}
  \vspace{-38mm}
  \label{fig:preprocess}
\end{wrapfigure}

As illustrated in \cref{fig:preprocess}, \texttt{Preprocess} randomly samples a timestep $t_\text{sample}^i$ for each instance $i$ (highlighted with colored contours).
This timestep determines when an instance shifts from being treated as a detection target to a tracking target. Using the ground truth $y_{1:T}^i$ and sampled index $t_\text{sample}^i$, it constructs, for each instance $i$ and frame $t$,
\begin{align*}
y^{\text{det},i}_{t} &= 
\begin{cases}
y_t^i & \text{if } t \leq t_\text{sample}^i \\
\varnothing & \text{otherwise}
\end{cases}\\
y^{\text{trk},i}_{t} &= 
\begin{cases}
y_t^i & \text{if } t > t_\text{sample}^i \\
\varnothing & \text{otherwise}
\end{cases}\\
\mathcal{M}_t^i &= 
\begin{cases}
y_t^i & \text{if } t \geq t_\text{sample}^i \\
\varnothing & \text{otherwise}
\end{cases}
\end{align*}
The set $\mathcal{A}_{t-1}$ is then constructed to contain vectors corresponding to foreground instances in $\mathcal{M}_{t-1}$, maintaining the one-to-one mapping between allocated vectors and instances.

Once preprocessing is complete, all subsequent operations, from applying History Marker to calculating losses, can be executed in parallel across frames.
The training loss decomposes into tracking loss $L^\text{trk}$ and detection loss $L^\text{det}$. For tracking, the one-to-one mapping between $y^\text{trk}_t$ and $\mathcal{A}_{t-1}$ enables direct loss computation.
For detection, where no predetermined mapping exists between $y^\text{det}_t$ and the detection queries $\mathcal{V}$, we employ the Hungarian algorithm~\citep{DETR} to obtain optimal assignments before computing the loss.

\section{Experiments}
\label{sec:experiment}

\subsection{Datasets}

\textbf{Training Datasets. }
We train \model{} on diverse public segmentation datasets encompassing both prompted and unprompted paradigms.
Specifically, we use COCO~\citep{COCO}, DAVIS 2017~\citep{DAVIS}, MOSE~\citep{MOSE}, SA-V~\citep{SAM2}, YouTube-VIS 2019 \& 2021~\citep{VIS}, and OVIS~\citep{OVIS}.
For datasets that include semantic labels~\citep{COCO, VIS, OVIS}, we provide an additional classification objective so the model learns category-aware segmentation.

\textbf{Evaluation Benchmarks and Metrics. }
For prompted tasks, we follow established semi-supervised VOS protocols on four benchmarks: DAVIS 2017, YouTube-VOS 2018 \& 2019~\citep{YTVOS}, and MOSE, which focuses on multi-object segmentation with complex instance interactions.
For DAVIS and MOSE, we report the standard $\mathcal{J} \& \mathcal{F}$ metric, which averages region similarity and contour accuracy.
For YouTube-VOS, we report $\mathcal{G}$, the average of $\mathcal{J} \& \mathcal{F}$ computed across both seen and unseen categories.
For unprompted tasks, we evaluate on VIS benchmarks: YouTube-VIS 2019 \& 2021 and OVIS, the latter featuring heavy occlusion and longer videos.
Following standard practice, we report Average Precision (AP).

\subsection{Training Details}
\label{sec:impl_detail}

A core design choice in \model{} is enabling \emph{parallel} training over frame sequences.
Our architecture processes all frames in a training clip concurrently while preserving the autoregressive \emph{training} objective over the output sequence.
The training process is divided into three stages, progressively increasing temporal complexity and data diversity.

\textbf{Stage 1 (Pseudo-video pretraining):}
We pretrain AUSM on COCO~\citep{COCO} using a pseudo-video augmentation strategy (COCO-pseudo).
Each image is transformed into a 3-frame sequence via random spatial augmentations following~\citep{VITA}.

\textbf{Stage 2 (Multi-source short-clip training):}  
This stage introduces real video data and train on 5-frame clips sampled from a mixture of COCO-pseudo, MOSE~\citep{MOSE}, SA-V~\citep{SAM2}, YouTube-VIS 2019 \& 2021~\citep{VIS}, and OVIS~\citep{OVIS}.

\textbf{Stage 3 (Long-clip adaptation):}
We fine-tune \model{} on 16-frame clips to strengthen long-range temporal modeling.
To reduce memory usage, we freeze the image backbone and update only the temporal modules and prediction heads.
This stage uses the same datasets as Stage 2, with the addition of the DAVIS 2017 training set~\citep{DAVIS}.

\begin{table}[t]
\centering
\caption{Quantitative results on Prompted and Unprompted Video Segmentation benchmarks. We report $\mathcal{J}\&\mathcal{F}$ for DAVIS and MOSE, $\mathcal{G}$ for YouTube-VOS, and AP for YouTube-VIS and OVIS. $\dag$ denotes methods additionally trained on private datasets. "\NA{}" indicates tasks that are architecturally incompatible with the models. "--" denotes tasks that are feasible to handle with the models but for which results are not reported.}
\vspace{2mm}
\resizebox{\textwidth}{!}{
\begin{tabular}{llccccccc}
    \toprule
    \multirow{2}{*}{Method} & \multirow{2}{*}{Backbone} &
    \multicolumn{4}{c}{\textit{Prompted}} &
    \multicolumn{3}{c}{\textit{Unprompted}} \\
    \cmidrule(l{4pt}r{4pt}){3-6} \cmidrule(l{4pt}r{4pt}){7-9}
    & & DAVIS & MOSE & YTVOS18 & YTVOS19 & YTVIS19 & YTVIS21 & OVIS \\
    \midrule
    \rowcolor{gray!10} \multicolumn{9}{l}{\textit{\textbf{\quad Task-specialized Models}}}\\
    Xmem~\cite{Xmem}         & ResNet50    & 86.2 & 57.6 & 85.7 & 85.5 & \NA & \NA & \NA \\
    DeAOT~\cite{DeAOT}       & ResNet50    & 85.2 & 59.4 & 86.0 & 85.9 & \NA & \NA & \NA \\
    DeAOT~\cite{DeAOT}       & Swin-B      & 86.2 & --   & 86.2 & 86.1 & \NA & \NA & \NA \\
    SAM2~\cite{SAM2}$^\dag$ & Hiera-B+    & 90.2 & 76.6 & --   & 88.6 & \NA & \NA & \NA \\
    SAM2~\cite{SAM2}$^\dag$ & Hiera-L     & 90.7 & 77.9 & --   & 89.3 & \NA & \NA & \NA \\
    UniRef++~\cite{UniRef++} & Swin-L      & 83.9 & 59.0 & 83.2 & 83.0 & \NA & \NA & \NA \\
    GenVIS~\cite{GenVIS}     & Swin-L      & \NA  & \NA  & \NA  & \NA  & 64.0 & 59.6 & 45.2 \\
    DVIS~\cite{DVIS}         & Swin-L      & \NA  & \NA  & \NA  & \NA  & 63.9 & 58.7 & 47.1 \\
    VISAGE~\cite{VISAGE}     & Swin-L      & \NA  & \NA  & \NA  & \NA  & 64.2 & 59.6 & 46.5 \\
    Video K-Net~\cite{VideoKNet} & Swin-B & \NA  & \NA  & \NA  & \NA  & 54.1 & --   & --   \\
    \rowcolor{gray!10} \multicolumn{9}{l}{\textit{\textbf{\quad Universal Offline Models}}}\\
    TarViS~\cite{TarViS} & Swin-T   & 82.8 & --   & --   & --   & --   & 50.9 & 34.0 \\
    TarViS~\cite{TarViS} & Swin-L   & 85.3 & --   & --   & --   & --   & 60.2 & 43.2 \\
    \rowcolor{gray!10} \multicolumn{9}{l}{\textit{\textbf{\quad Universal Streaming Models}}}\\
    UNINEXT~\cite{UNINEXT}   & ResNet50   & 74.5 & --   & 77.0 & --   & 53.0 & --   & 34.0 \\
    UNINEXT~\cite{UNINEXT}   & ConvNeXt-L & \underline{77.2} & --   & 78.1 & --   & \textbf{64.3} & --   & 41.1 \\
    UniVS~\cite{UniVS}       & Swin-T     & 71.7 & --   & 70.3 & --   & 52.4 & 51.6 & 33.0 \\
    UniVS~\cite{UniVS}       & Swin-B     & 75.0 & --   & 70.9 & --   & 57.8 & 56.5 & 39.0 \\
    UniVS~\cite{UniVS}       & Swin-L     & 76.2 & --   & 71.5 & --   & 60.0 & \underline{57.9} & \underline{41.7} \\
    \rowcolor{nvgreen!20} AUSM                    & Swin-T     & 76.4 & \underline{58.8} & \underline{79.5} & \underline{78.3} & 54.9 & 52.1 & 39.4 \\
    \rowcolor{nvgreen!20} AUSM                    & Swin-B     & \textbf{81.6} & \textbf{62.1} & \textbf{80.2} & \textbf{79.1} & \underline{62.6} & \textbf{58.6} & \textbf{45.5} \\
    \bottomrule
\end{tabular}
}
\label{tab:main_result}
\end{table}

\subsection{Main Results}
Table~\ref{tab:main_result} reports AUSM's performance across a range of prompted and unprompted video segmentation benchmarks.
We split the table into two categories: (1) specialized methods tailored to a single setting -- prompted or unprompted -- and (2) universal frameworks that support both within one architecture.

All results of AUSM are obtained with a \emph{single model} trained using our joint learning framework, without task-specific fine-tuning.
We report AUSM with Swin-T and Swin-B backbones~\citep{Swin}.

\textbf{Prompted Video Segmentation. }
A leading specialized method for prompted video segmentation is SAM2~\citep{SAM2}, a memory-based mask-propagation approach built on STM~\citep{STM}, where each object is processed independently using a dedicated memory buffer.
With additional private data, SAM2 attains strong performance across standard benchmarks.
In contrast, \model{} processes multiple objects jointly in a single forward pass and operates without explicit memory buffers.
Among online universal frameworks, AUSM performs strongly, demonstrating the benefits of our autoregressive formulation.
Notably, relative to UniVS~\citep{UniVS}, we surpass its Swin-L variant on YouTube-VOS 2018 by $+8.7$ (in $\mathcal{G}$) despite using a smaller Swin-B backbone.
This highlights \model{} as a unified, memory-efficient alternative that balances generality and scalability.

\textbf{Unprompted Video Segmentation. }
Specialized VIS models achieve strong results across benchmarks, but their decoupled designs, often tailored for object-query propagation~\citep{GenVIS} or short-term tracking during training~\citep{VISAGE}, limit extensibility to prompted tasks.
Despite being universal, \model{} is competitive with these specialized approaches, capturing complex object dynamics without task-specific architectural constraints.
This underscores the strength of our autoregressive formulation in handling unprompted segmentation while preserving compatibility with prompted settings.
Notably, \model{} attains the highest OVIS score among universal models, a dataset characterized by heavy occlusion and long-range interactions.

\subsection{Ablation Studies}
\begin{figure*}[t]
    \centering
    \begin{minipage}{0.48\textwidth}
        \centering
        \includegraphics[width=\linewidth]{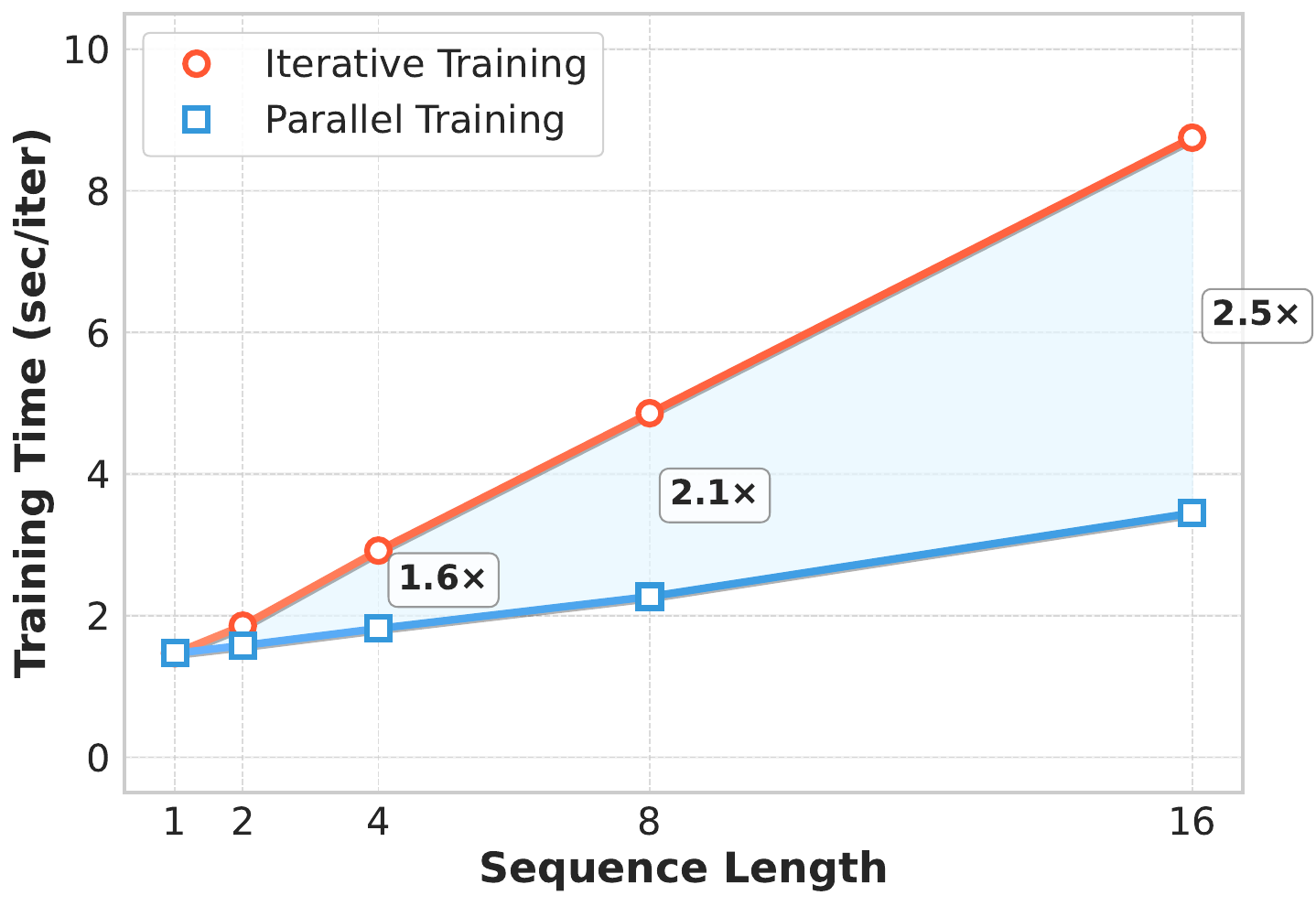}
        \vspace{-4mm} 
        \caption{Comparison of training time (sec/iter) between iterative and parallel training approaches across different sequence lengths.}
        \label{fig:abb-training-time}
    \end{minipage}
    \hfill
    \begin{minipage}{0.48\textwidth}
        \centering
        \includegraphics[width=\linewidth]{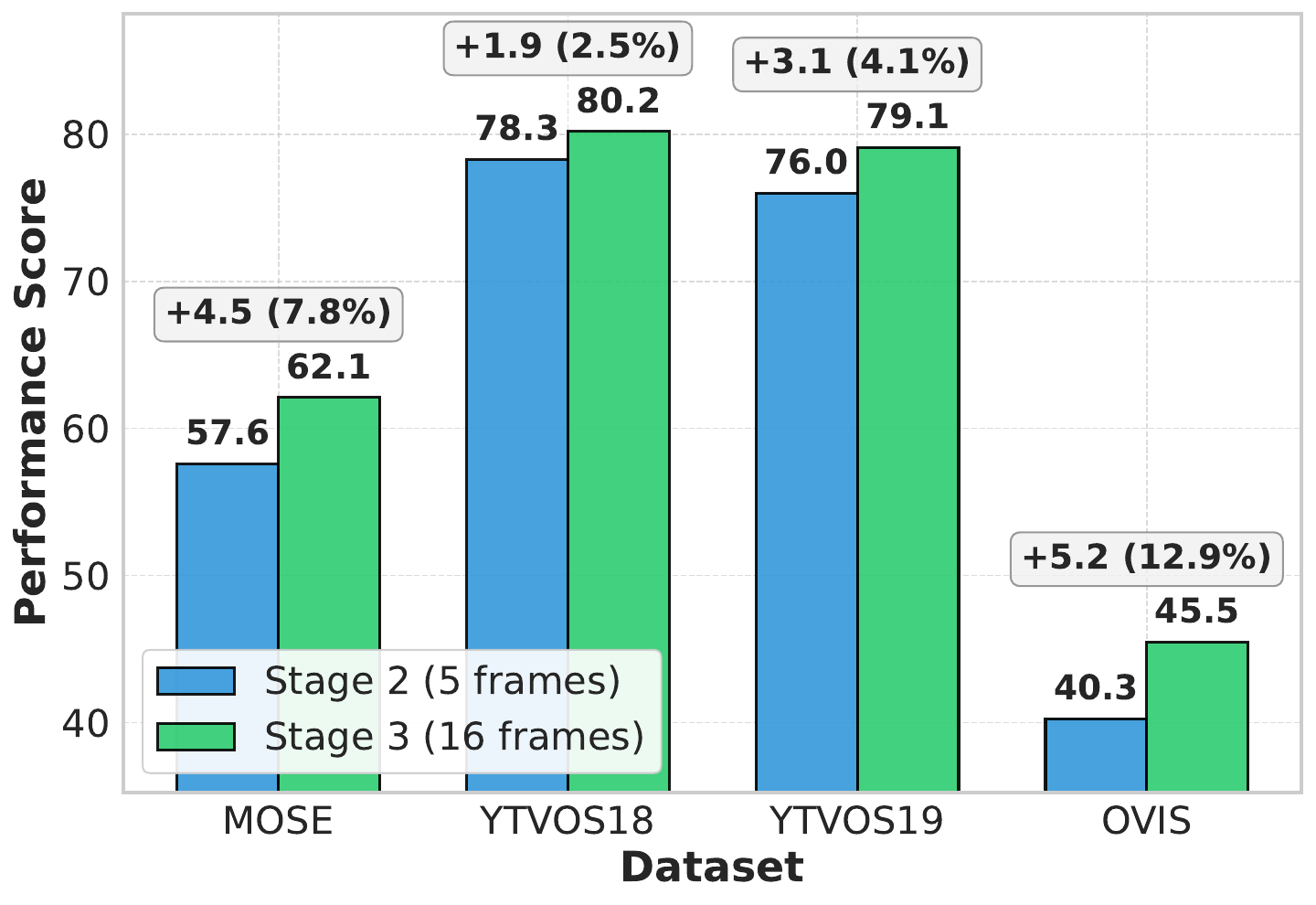}
        \vspace{-4mm} 
        \caption{Performance comparison between Stage 2 (5-frame) and Stage 3 (16-frame) training across four benchmark datasets.}
        \label{fig:abb-stage-comp}
    \end{minipage}
\vspace{-1mm}
\end{figure*}

\textbf{Parallel vs.\ Iterative Training Efficiency.}
To quantify the benefit of parallel training, we measure training time per iteration (s/iter) over sequence lengths of 1, 2, 4, 8, and 16 using the Swin-B backbone.
As shown in~\cref{fig:abb-training-time}, our parallel approach scales substantially better with increasing sequence length compared to the iterative baseline.
Whereas the iterative method grows from $1.47$s to $8.75$s per iteration, our parallel approach increases only from $1.47$s to $3.45$s at length 16.
This yields a $2.5\times$ speedup at sequence length 16, with larger gains expected at longer horizons.
These results highlight the efficiency and scalability of our parallel training pipeline, making it especially well-suited for learning from long video sequences where temporal modeling is critical.

\textbf{Effect of Training with Longer Sequences.}
We compare Stage~2 (5-frame clips) and Stage~3 (16-frame clips) to assess our long-clip adaptation strategy.
As shown in~\cref{fig:abb-stage-comp}, the transition to longer sequences consistently improves performance across all datasets.
The largest gains are on MOSE ($+4.52$) and OVIS ($+5.2$), indicating that longer temporal context improves modeling of complex dynamics and appearance changes.
These improvements are achieved without any explicit memory buffer (\eg, FIFO-style spatio-temporal caches); instead, our History Comperssor enables efficient long-term reasoning with constant memory.

\begin{wraptable}{r}{0.45\textwidth}
    \vspace{-5mm}
    \caption{
    Effect of varying the foreground threshold during inference on Unprompted Video Segmentation performance.
    }
    \centering
    \vspace{-2mm}
    \begin{tabular}{c|ccc}
        \toprule
        Thres. & YTVIS19 & YTVIS21 & OVIS \\
        \midrule
        0.3 & 62.4 & 57.8 & 44.5\\
        0.4 & \textbf{62.6} & \underline{58.1} & 45.1\\
        0.5 & \textbf{62.6} & \textbf{58.6} & 45.5\\
        0.6 & \underline{61.8} & \underline{58.1} & \underline{46.4}\\
        0.7 & 61.6 & 57.9 & \textbf{46.5}\\
        \bottomrule
    \end{tabular}
    \label{tab:fg-threshold}
    \vspace{-4mm}
\end{wraptable}
\textbf{Effect of Foreground Threshold at Inference.}
In unprompted inference, new objects enter the buffer via $\texttt{filter\_fg}(\hat{y}_t^{\text{det}})$, which selects confident detections using a foreground-probability threshold.
As shown in~\cref{tab:fg-threshold}, performance is relatively robust across thresholds from $0.3$ to $0.7$.
While OVIS benefits from a higher threshold (up to $46.5$ AP at $0.7$), YouTube-VIS performs best around $0.4$--$0.5$.
We therefore use a fixed threshold of $0.5$ for all benchmarks to ensure consistency and avoid dataset-specific tuning.

\textbf{Scaling Inference Compute in \model{}.}
Recent work in language modeling shows that increasing inference-time computation can improve accuracy~\citep{CoT}; even simple input-repetition strategies~\citep{Echo, JRT} help across architectures~\citep{Transformer, Mamba}.
Inspired by this, we scale inference compute for AUSM in \emph{both images and videos} by constructing repeated sequences that allow iterative refinement.

\begin{wraptable}{r}{0.38\textwidth}
    \vspace{-5mm}
    \caption{Results measured using Swin-B backbone. We relegate more experimental details in the supplementary.}
    \centering
    \begin{tabular}{ccc}
    \toprule
    \# Repetition   & COCO  & YTVIS19   \\
    \midrule
    $\times 1$      & 34.2  & 62.6      \\
    $\times 2$      & 34.9  & 63.3      \\
    $\times 3$      & 35.0  & 63.5      \\
    \bottomrule
    \end{tabular}
    \vspace{-2mm}
    \label{tab:scaling}
\end{wraptable}
For a single image $\mathcal{I}$, we construct a pseudo-video sequence $\mathcal{I}^\text{aug} = (\mathcal{I}_1, \ldots, \mathcal{I}_T)$ where $\forall t \in \{1,\ldots,T\}, \mathcal{I}_t = \mathcal{I}$.
Processing this sequence encourages the model to refine predictions by revisiting uncertain regions and leveraging object interactions over repetitions.
Similarly, we extend this concept to video sequences as well.
Given a video $(\mathcal{I}_1, \ldots, \mathcal{I}_T)$, we construct an augmented sequence by repeating the video frames as:
\begin{equation*}
    \mathcal{I}^\text{aug} = (\mathcal{I}_1, \ldots, \mathcal{I}_T, \mathcal{I}_{T-1}, \ldots, \mathcal{I}_1, \mathcal{I}_2, \ldots, \mathcal{I}_T)
\end{equation*}
When processing this augmented sequence, we take predictions from only the final $T$ frames as the refined output.

As shown in \cref{tab:scaling}, scaling inference compute improves performance.
While the primary focus of AUSM is unifying video segmentation tasks within one framework, these results show it can also benefit from test-time compute scaling for further gains in both images and videos.
\section{Related Work}
\label{sec:related_work}

\textbf{Unprompted Video Segmentation. }
A key distinction among existing models is whether they condition on past predictions $\hat{y}_{<t}$ to generate the current output during inference. Conventional online tracking-by-detection approaches completely exclude $\hat{y}_{<t}$ in their model design, treating each visual feature independently from previous predictions~\citep{MinVIS,IDOL,VISAGE,CTVIS}.
While these methods benefit from parallelizable computation during training, they often require external memory banks to maintain temporal consistency at inference time.

In contrast, models such as GenVIS~\citep{GenVIS} adopt query propagation by conditioning on prior predictions $\hat{y}_{<t}$ through object-level vector representations.
Although this formulation enhances temporal coherence, object vectorization significantly degrades the granularity of mask predictions~\citep{VISAGE}.
RoCoVIS~\citep{RoCoVIS} addresses this by introducing instance mask propagation, which greatly improves mask quality.
However, this design inherently breaks parallelism, as predictions must be generated sequentially, leading to reduced training efficiency.

Another line of work follows an offline paradigm, where the entire video $\mathcal{I}_{1:T}$ is available in advance during inference~\citep{VisTR,IFC,VITA}.
While this setting enables models to access long-range context, such models are typically not conditioned on intermediate outputs $\hat{y}$ and therefore cannot refine predictions based on prior object states.
As a result, despite high computational cost, these models often underperform recurrent methods~\citep{IDOL} that explicitly leverage temporal feedback.

\textbf{Prompted Video Segmentation. }
Prompted Video Segmentation focuses on segmenting and tracking an object throughout a video, given a specified target in the first frame, without requiring any class labels.
Unlike Unprompted Segmentation tasks that involve discovering and classifying all objects, this task aims to track a user-specified target.
Prompted Video Segmentation is highly practical, particularly for interactive and general-purpose applications.

One of the most influential paradigms is the Space-Time Memory Network (STM)~\citep{STM}, which has inspired a wide range of follow-up methods.
STM maintains a memory bank consisting of past frames and their corresponding masks, and performs dense matching between the current frame and the stored memory to retrieve relevant information for accurate mask propagation.
Building on this, XMem~\citep{Xmem} introduces enhanced memory mechanisms for improved long-term tracking.
More recently, SAM2~\citep{SAM2}, building upon SAM~\citep{SAM}, supports more flexible input representations such as points and boxes, and also introduces large-scale datasets to enable broader evaluation and training.

Another line of research explores hierarchical propagation using transformer-based architectures~\citep{AOT, DeAOT}.
These methods gradually propagate identity information from past frames to the current frame through a hierarchical structure.
It supports multi-object processing in a unified manner, unlike STM-based approaches that typically handle each object separately.

\textbf{Universal Video Segmentation. }
The emergence of transformer-based detection and segmentation architectures~\citep{DETR, MaxDeepLab, Mask2Former} has facilitated early attempts to unify multiple tasks within unprompted video segmentation (\eg VIS, VPS, and VSS) under a single framework~\citep{TubeFormerDeepLab, VideoKNet}.
In parallel, similar efforts have been made in the prompted setting, aiming to jointly handle VOS and referring VOS within a shared architecture~\citep{UniRef, UniRef++}.

Recent research has progressed toward bridging the gap between Unprompted and Prompted Video Segmentation.
These efforts aim to develop universal models that can support both interaction-driven and fully automatic scenarios within a unified framework, thereby reducing reliance on task-specific designs and enabling broader applicability across diverse video understanding tasks.
An offline method, TarViS~\citep{TarViS}, represents the first attempt to jointly model both settings by encoding task-specific targets as a set of queries.
UNINEXT~\citep{UNINEXT} further introduces a prompt-guided object discovery and retrieval paradigm.
UniVS~\citep{UniVS} leverages prompts as queries by treating predicted masks from previous frames as visual prompts for the current frame.

\textbf{State Space Models.}
State Space Models (SSMs)~\citep{S4, S4D, Mamba} have recently emerged as a promising alternative to Transformer~\citep{Transformer} for sequence modeling.
While Transformer requires storing all tokens in key-value caches, SSMs compress all history into a single state.
This property gives SSMs constant computational complexity and memory requirements during inference, providing significant advantages for modeling long sequences compared to Transformers' linearly growing complexity and memory.

\section{Discussion}
\label{sec:discussion}

Our reformulation of video segmentation as an autoregressive modeling problem draws a direct conceptual connection to language modeling and opens several promising research directions.
One central direction is handling long sequences, with two complementary areas: (i) retrieval for long contexts (e.g., needle-in-a-haystack (NIAH) evaluations)~\citep{RULER} and (ii) length extrapolation~\citep{ALiBi, AttentionSink}.
Our model demonstrates strong performance and is capable of processing arbitrarily long streams.
However, similar to LLMs, we observe performance degradation on extremely long sequences.
Recent long-sequence techniques from language modeling could be adapted to video to maintain quality and extend the effective context beyond the training sequence length.

Another promising direction is extending AUSM to additional video perception tasks.
While this work focuses on both prompted and unprompted video segmentation (VOS and VIS), other tasks can be integrated with minimal modifications.
For example:
\begin{itemize}
    \item Object tracking~\citep{OnlineObjectTracking} and multi-object tracking~\citep{MOT16}: convert bounding boxes into mask prompts by filling corresponding regions with masks so they fit the same segmentation interface.
    \item Taking the same approach as bounding boxes, other prompt styles (\emph{e.g.,} scribbles, points) can be unified by converting annotations into mask signals.
    \item Referring video object segmentation~\citep{RVOS}: initialize the History Compressor's state with text embeddings (e.g., from a frozen text encoder) corresponding to the language prompt.
\end{itemize}

We expect incorporating more tasks and data to yield further performance gains, and training on longer videos should also improve long-context modeling.
Although our experiments train on clips up to 16 frames due to memory constraints, the framework is designed to scale and will benefit from next-generation hardware.

\textbf{Limitations.}
While \model{} attains strong results -- and among universal/online methods, state-of-the-art performance on unprompted VIS (\emph{e.g.,} OVIS) -- we observe a modest gap on prompted VOS compared to specialized, memory-heavy systems.
We believe these results are primarily from our architectural choice: most modules of AUSM take coarse frame features (\emph{e.g.} stride of 8), which saves memory than finer features (\emph{e.g.,} stride of 4) and is better suited for object-level understanding, but marginally worse in capturing details.
Future work could be a new video-specialized backbone to temporal modeling (\emph{e.g.,} reducing frame-independent layers while strengthening frame-dependent modules such as History Compressor/Decoder and prompt conditioning) which may close the VOS gap without sacrificing unprompted performance.

\section{Conclusion}
\label{sec:conclusion}

We introduce \model{} as a step toward a unified, scalable, and general-purpose formulation of video segmentation.
Leveraging the connection between language modeling and video segmentation, \model{} supports parallel training and performs temporal modeling without explicit memory buffers, enabling efficient long-range reasoning with constant-memory inference.
Across diverse benchmarks, our experiments show that autoregressive modeling provides a practical and scalable solution for video segmentation.
We hope this perspective serves as both a strong baseline and a useful direction for future research in video perception.

\textbf{Acknowledgements. } We thank De-An Huang for feedback and helpful discussions.

\clearpage
%%%%%%%%%%%%%%%%%%%%%%%%%%%%%%%%%%%%%%%%%%%%%%%%%%%%%%%%%%%%

\appendix

\section{Implementation Details}
\label{sec:supp_imple_details}

We further summarize key architectural configurations and experimental setups used throughout the training and evaluation of \model{}.

\textbf{Model Configuration. }
The number of object queries for detection $N^\text{det}$ and ID vectors $N^\text{id}$ are both set to 100. The \texttt{History Compressor} module consists of 6 layers, and the following Transformer decoder contains 6 layers. For spatio-temporal fusion in \texttt{History Compressor}, we use the $1/8$ resolution feature map from the Swin backbone~\citep{Swin}, with a fusion feature dimension of 256.

Our implementation does not use vision-language supervision (e.g., CLIP~\citep{CLIP} or other pretrained image-text models); we instead employ dataset-specific classification heads. During training, each head is conditionally selected according to the dataset from which the input sample originates.

\textbf{Training Setup. }
All experiments are conducted using 16 NVIDIA A100 GPUs with a batch size of 16 and an initial learning rate of $1 \times 10^{-4}$, optimized using AdamW~\citep{AdamW} across all stages.

We begin the training process with pseudo-video training on the COCO dataset~\citep{COCO}, using image instance segmentation pretrained Mask2Former~\citep{Mask2Former} weights as initialization. This stage is trained for 20 epochs, corresponding to 147{,}500 iterations.

Next, we perform multi-source short-clip training using 5-frame video segments drawn from multiple datasets. This stage runs for 32{,}000 iterations and serves to adapt the model to short-term temporal dynamics and diverse visual domains.  

Finally, we apply long-clip adaptation by fine-tuning the model on 16-frame clips for 40{,}000 iterations, enabling the model to better capture long-range temporal dependencies.  

\textbf{Inference Setup. }
Given a frame, we set the resolution of the shorter side to $1024$.
For Unprompted Video Segmentation, we apply a fixed foreground probability threshold of 0.5 to select confident detection predictions. Additionally, we perform top-$k$ selection, retaining the top 10 instances per frame for YouTube-VIS 2019/2021~\citep{VIS} and the top 20 for OVIS~\citep{OVIS}, to measure benchmark evaluation. These are the only post-processing steps employed; no other heuristics are introduced.

In contrast, prompted video segmentation involves no thresholding, filtering, or post-processing. Inference in this setting is fully model-driven, without reliance on manually designed components.

\textbf{Implementation Details for Inference Scaling Experiments.}
We present additional methodological considerations for the scaling inference compute experiments.
Due to the dense number of object appearing in the COCO validation set, we set dataset-specific foreground confidence thresholds to mitigate propagation of erroneous detections.
Specifically, we use foreground confidence thresholds of 0.9 and 0.5 for COCO and YTVIS datasets, respectively.
This stringent threshold for COCO results in a relatively modest performance metric ($34.2$ AP), as a considerable proportion of valid but lower-confidence detections are excluded from evaluation.

We further explore an alternative data augmentation strategy beyond simple image repetition.
This approach involves strategic spatial decomposition of the input image into four overlapping quadrants: upper-left ($\mathcal{I}^\text{UL}$), upper-right ($\mathcal{I}^\text{UR}$), bottom-right ($\mathcal{I}^\text{BR}$), and bottom-left ($\mathcal{I}^\text{BL}$).
Each quadrant maintains substantial spatial redundancy, with dimensions set to $90\%$ of the original image size. 

The resulting augmented sequence is formulated as:
\begin{equation}
\mathcal{I}^\text{aug} = (\mathcal{I}, \mathcal{I}^\text{UL}, \mathcal{I}^\text{UR}, \mathcal{I}^\text{BR}, \mathcal{I}^\text{BL}, \mathcal{I})
\end{equation}

This configuration enables the model to systematically traverse local regions while maintaining global context through the inclusion of the full image at sequence boundaries.
Empirical evaluation demonstrates that this spatial traversal strategy yields statistically significant performance improvements, increasing mAP from $34.2$ to $35.9$ on the COCO validation set.

\section{Qualitative Results}
\label{sec:qualitative}
\begin{figure}[t]
\begin{center}
\includegraphics[width=\linewidth]{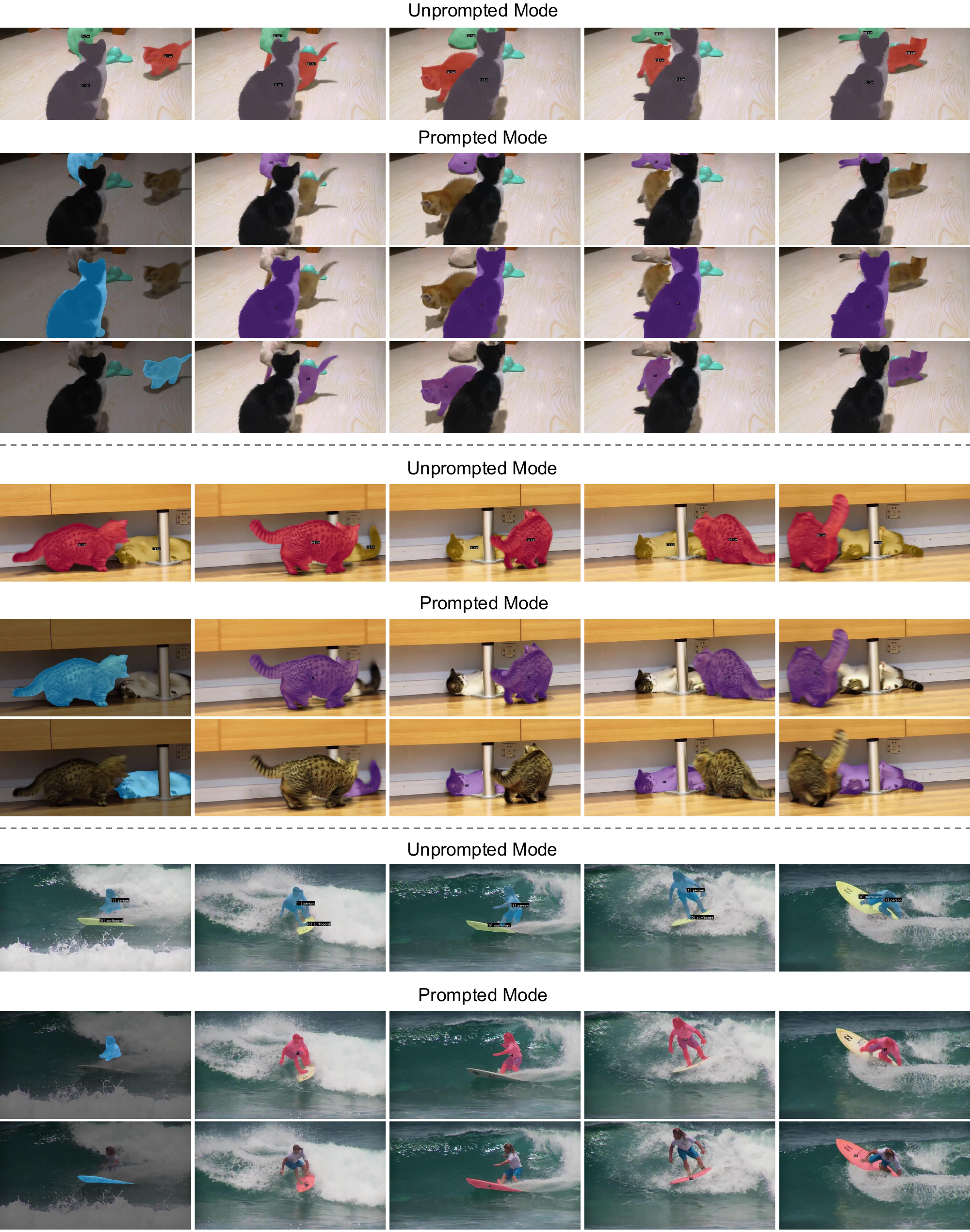}
\end{center}
\caption{
Qualitative comparison between unprompted and prompted video segmentation results using a single \model{} model. 
In the unprompted mode, the model autonomously discovers, segments, and classifies objects in the scene without any external guidance. 
In the prompted mode, it tracks only the object specified by an initial mask in the first frame. 
These examples demonstrate the unified capability of \model{} to seamlessly support both modes within a single framework.
}
\label{fig:qualitative1}
\end{figure}

\begin{figure}[t]
\begin{center}
\includegraphics[width=\linewidth]{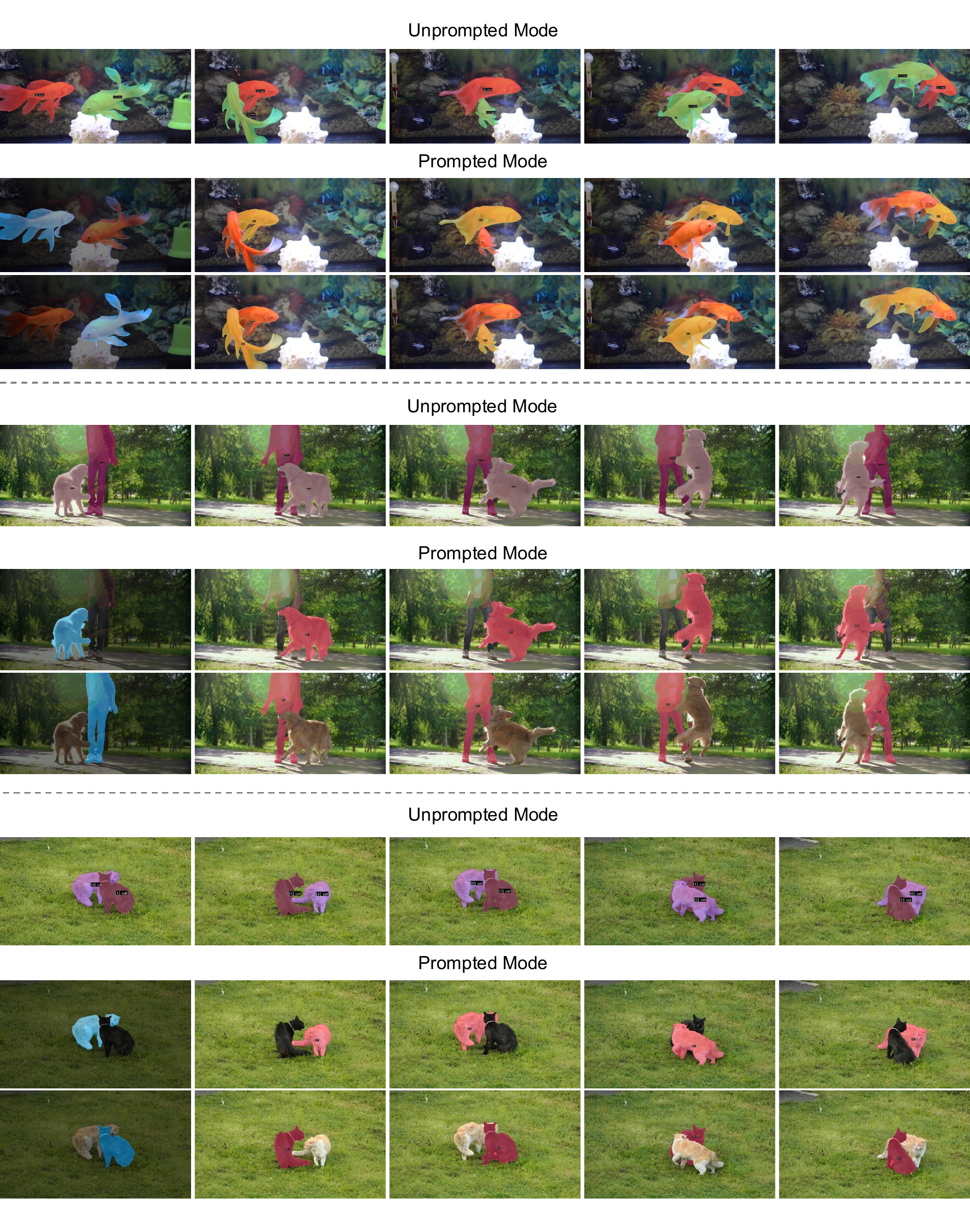}
\end{center}
\caption{
Qualitative comparison between unprompted and prompted video segmentation results using a single \model{} model. 
In the unprompted mode, the model autonomously discovers, segments, and classifies objects in the scene without any external guidance. 
In the prompted mode, it tracks only the object specified by an initial mask in the first frame. 
These examples demonstrate the unified capability of \model{} to seamlessly support both modes within a single framework.
}
\label{fig:qualitative2}
\end{figure}

In~\cref{fig:qualitative1} and~\cref{fig:qualitative2}, we present qualitative results to illustrate the capability of \model{} in both prompted and unprompted video segmentation settings. Notably, a single trained model is used for all results without any task-specific tuning or architectural changes. This underscores the unified nature of \model{}, which can seamlessly switch between prompted and unprompted modes at inference time.

For prompted segmentation, we visualize the model’s ability to accurately segment target objects given masks in the keyframe. For unprompted segmentation, we show how the model discovers and tracks multiple objects throughout a video without external guidance. These results demonstrate that \model{} effectively handles both interaction-based and autonomous video understanding scenarios within a unified architecture. 


\clearpage
\setcitestyle{numbers}
\bibliographystyle{plainnat}
\bibliography{main}

\end{document}